\documentclass{article} 

\usepackage{subcaption}
\usepackage{graphicx}

\usepackage{algorithm}
\usepackage{algpseudocode}

\usepackage{collas2026_conference,times}
\usepackage{easyReview}

\usepackage{appendix}

\usepackage{amsmath,amsfonts,bm}

\def\eqref#1{equation~\ref{#1}}

\def\1{\bm{1}}

\DeclareMathAlphabet{\mathsfit}{\encodingdefault}{\sfdefault}{m}{sl}
\SetMathAlphabet{\mathsfit}{bold}{\encodingdefault}{\sfdefault}{bx}{n}

\usepackage{hyperref}

\hypersetup{
    colorlinks=true,
    linkcolor=red,
    filecolor=magenta,
    urlcolor=blue,
    citecolor=purple,
    pdftitle={Overleaf Example},
    pdfpagemode=FullScreen,
    }

\title{Generative Replay for Equilibrium Propagation: Toward Power and Memory-Efficient Continual Learning in Hopfield Networks}

\author{Victor Cook, Gilles Sassatelli, Thierry Gil, Mohamed Watfa\\
LIRMM\\
Univ. Montpellier / CNRS, Montpellier\\
France \\
\texttt{\{victor.cook,gilles.sassatelli,thierry.gil,mohamed.watfa\}@lirmm.fr} \\
\And 
Marina Reyboz  \\
CEA List \\
Univ. Grenoble Alpes, Grenoble \\
France \\
\texttt{marina.reyboz@cea.fr}
}

\preprintcopy 

\begin{document}

\maketitle

\begin{abstract}

Equilibrium Propagation (EqProp) offers a biologically grounded and hardware-aligned alternative to backpropagation for training Hopfield Networks. However, its behavior in continual learning settings remains largely unexplored.

This work confronts catastrophic forgetting in EqProp-trained Hopfield Networks from the inside out, eliminating the need for external data buffers or architectural overhead that standard Experience Replay \textbf{(ER)} and Generative Replay \textbf{(Gen-Replay)} inevitably require. We introduce an Energy-Based Generative Replay strategy \textbf{(EB-Gen)} which exploits the intrinsic generative dynamics of Hopfield Networks, allowing a single model to act as both classifier and replay samples generator. Samples capturing past task knowledge are generated by relaxing a frozen copy of the network, enabling rehearsal solely through internally generated pseudo-data.

We evaluate the proposed method in the Class-Incremental setting, using the split versions of the MNIST and CIFAR10 benchmarks and compare it with FineTune, Experience Replay, Generative Replay, Sleep-Like Replay Consolidation (SRC) and Joint training. Our results show that EB-Gen significantly improves stability while preserving plasticity, achieving a 75\% and 64\% average accuracy on Split MNIST and Split CIFAR10 respectively. These findings show that Equilibrium Propagation can be adapted to alleviate catastrophic forgetting without forgoing the traits which made it compelling in the first place.
\end{abstract}

\section{Introduction}
Continual learning (CL) studies how a model can acquire new knowledge from a
non-stationary data stream while retaining previously learned information
~(\cite{lin1992self,parisi201954,delange2023}).
Standard neural networks trained with gradient descent fail in this regime due
to \emph{catastrophic forgetting}, where learning new tasks rapidly degrades
performance on earlier ones~(\cite{CatastrophicInterferenceConnectionist1989}).

Prior work has explored a wide variety of methods such as parameter
regularization~(\cite{kirkpatrick2017ewc,zenkeContinualLearningSynaptic2017}),
dynamic or task-expanding
architectures~(\cite{rusu2022progressiveneuralnetworks}), and replay-based
strategies~(\cite{lopezpaz2017gem,shinContinualLearningDeep2017}). While effective, these approaches often rely on explicit memory buffers, auxiliary models, or architectural growth—requirements that conflict with the constraints of resource-limited environments.

Energy-Based Models (EBMs) offer a promising alternative for such settings. By framing inference as energy minimization, they align naturally with analog hardware implementations, where power efficiency is critical. Moreover, ~\cite{liEnergyBasedModelsContinual2022} demonstrated that EBMs exhibit inherent resistance to forgetting due to their reliance on an energy function--an unnormalized probability distribution over data. However, their results were obtained using backpropagation, a globally optimized rule, leaving open the question of how EBMs behave under local learning regimes.

Equilibrium Propagation (EqProp), introduced by Scellier \& Bengio (2017), provides a compelling framework to address this gap. Designed explicitly for energy-based networks, EqProp replaces backpropagation with a local learning rule derived from the system's relaxation dynamics. By comparing two equilibrium states—obtained via a free relaxation and a weakly nudged one—it enables training without needing any additional mechanisms. When applied to Hopfield Networks, EqProp unifies inference and learning under the same energy-minimization dynamics, allowing the network to sculpt its attractor landscape without auxiliary components. 
While EqProp has been extensively studied as a training paradigm for EBMs (e.g., \cite{ernoultUpdatesEquilibriumProp2019,laborieuxScalingEquilibriumPropagation2021}), its potential for continual learning remains largely unexplored.

Recently, \cite{kuboLifelongLearningEquilibrium2025} introduces Sleep-Like Replay Consolidation (SRC), an EqProp-based method alleviating catastrophic forgetting in EqProp-trained Hopfield Networks by adding a Spike-Timing Dependant Plasticity - based knowledge consolidation phase before learning each new task. Our work investigates incremental learning in a more general setting, without employing spiking neural networks.



We demonstrate that Hopfield Networks trained with EqProp can be readily adapted to enable generative replay. Since learned tasks are represented as stable attractors, relaxing the network from unconstrained or weakly perturbed states consistently yields coherent configurations that capture previously acquired structures.
We exploit this property to introduce \emph{Energy-Based Generative Replay} (EB-Gen in short), a continual learning strategy in which a frozen copy of the currently training network generates pseudo-examples by relaxation and interleaves them with current task data. 
This approach allows replay to emerge naturally from the same dynamics used for inference and learning, requiring no stored data, auxiliary components, or alterations to the underlying model topology.

The contributions of this work are as follows:

\begin{itemize}

    \item We show that EqProp does not naturally train Hopfield Networks to model the input distribution, and we propose a modification to the EqProp algorithm that effectively addresses this limitation,

    \item Building on this insight, we introduce EB-Gen, an Energy-Based Generative replay method that leverages the intrinsic generative dynamics of EqProp-trained Hopfield Networks without the need for stored data or generative models.

    \item We empirically evaluate the proposed approach on the Split-MNIST - CIFAR10
    benchmarks, comparing them against FineTune, ER, SRC and standard Generative Replay to
    quantify stability--plasticity trade-offs.

\end{itemize}

The remainder of this paper is organized as follows. Section~\ref{sec:related_works} reviews related work. Section~\ref{sec:method} describes the models, datasets, and experimental methodology. Section~\ref{sec:ebgenreplay} details the proposed EB-Gen approach. Experimental results are presented in Section~\ref{sec:results} and further analyzed in Section~\ref{sec:analysis}.

\section{Related Works}
\label{sec:related_works}
Catastrophic forgetting is a phenomenon which affects all neural networks when they are sequentially trained on non-\emph{identically and independently distributed} (i.i.d) data~(\cite{CatastrophicInterferenceConnectionist1989}). This scenario might be particularly prevalent in resource-constrained environments, which makes it a particularly relevant setting for testing EqProp-trained Hopfield Networks as their main appeal is their amenability for low power analog hardware implementations.


\paragraph{Replay-based Continual Learning methods}

Numerous continual learning methods have been proposed. In this work,  we concentrate on replay-based methods. Those methods mitigate forgetting by interleaving samples that represent previously learned data with examples from the current task during training. These replay samples are drawn from a reduced representation of previous tasks, which may manifest as a sparse memory buffer or a learned generative distribution. Experience Replay (ER), first introduced in ~\cite{lin1992self}, uses a simple buffer to store some of its training examples for replaying them later on. Generative Replay methods, use dedicated models to generate data inputs from former tasks. The generative models can be VAEs (\cite{venGenerativeReplayFeedback2019}), GANs (\cite{shinContinualLearningDeep2017}), diffusion models (\cite{gaoDDGRContinualLearning2023}) etc. Although most methods use distinct models for generation and classification, some methods use a hybrid model to simplify the training process and share learned representations (\cite{venGenerativeReplayFeedback2019}). For more complex vision benchmarks, \cite{liuGenerativeFeatureReplay2020,stoychevLatentGenerativeReplay2023,jeongReplayingRealisticLatent2025} train their generators on extracted features rather than raw inputs.
Early foundational work \cite{CatastrophicForgettingRehearsal,ansSequentialLearningDistributed2004a} focused on more biologically -inspired and -plausible methods often grouped under the name "pseudo-rehearsal".

\paragraph{Hopfield Networks}

Hopfield networks, first introduced by \cite{j.j.hopfieldNeuralNetworksPhysical1982}, are a class of recurrent neural networks that serve as associative memories, enabling the storage and retrieval of patterns. They comprise a set of neurons linked through bidirectional connections and are designed to converge toward stable states that encode stored patterns. Follow up works upgraded them from discrete to continuous state spaces~\cite{hopfieldNeuronsGradedResponse1984}.

\paragraph{Learning Rule}

In~\cite{j.j.hopfieldNeuralNetworksPhysical1982},
Hopfield used Hebbian learning to train their models on toy problems. \cite{hintonOptimalPerceptualInference} introduced Contrastive Hebbian Learning
to enable training Hopfield Networks for supervised tasks which require non-linear modeling capabilities. Eventually,
\cite{scellierEquilibriumPropagationBridging2017a} inntroduced Equilibrium Propagation with the goal of bridging Hopfield networks and standard neural networks trained via backpropagation.

\paragraph{Up-Scaling Hopfield Networks} Hopfield Networks are notoriously difficult to train on larger scale problems (even when using EqProp). In reaction, a part of the literature focused on improving the pattern storage capacity of Hebbian-trained hopfield networks \cite{krotovDenseAssociativeMemory2016,ramsauerHopfieldNetworksAll2021} while other works focused on improving EqProp by implementing it on the more potent CNN network topology ~(\cite{ernoultUpdatesEquilibriumProp2019}) and by improving training performance of EqProp by means of bettering gradient approximation \cite{laborieuxScalingEquilibriumPropagation2021,laborieuxHolomorphicEquilibriumPropagation2022}.
Although these works achieved significant performance improvements on benchmarks like CIFAR10, they often come at the expense of changes to the model or the learning rule that reduces their hardware friendliness. 

\paragraph{Training Hopfield Networks on Analog Hardware}
Analog hardware implementations of Equilibrium Propagation (EP) have advanced from theoretical proposals to practical demonstrations, leveraging the algorithm’s inherent compatibility with energy-efficient, parallel architectures. In their seminal paper, \cite{scellierEquilibriumPropagationBridging2017a} noted EqProp's potential for analog hardware due to local learning rules. Subsequent works validated this vision: \cite{zoppoEquilibriumPropagationMemristorBased2020} and \cite{ohMemristorCrossbarCircuits2023}. implemented EP on memristor crossbars, though weight updates were still computed externally. \cite{kendallTrainingEndtoEndAnalog2020a} extended these efforts to end-to-end analog training. Finally, \cite{watfaLowRankEquilibriumPropagation2025} proposed a low-rank EP variant that computes gradients on-device, eliminating the need for external digital computation and enabling fully analog training.

\paragraph{Continual Learning and EBMs}

The problem of training EBMs in a continual learning setting has already been addressed a few times in the literature. Notably,
\cite{liEnergyBasedModelsContinual2022} showed that Neural EBMs trained with contrastive divergence were much better at alleviating catastrophic forgetting than conventional neural
networks (designated as "Softmax-Based Classifiers").

\cite{singhWakeSleepEnergyBased2024} and
\cite{kuboLifelongLearningEquilibrium2025} both implemented methods based on a "wake-sleep" cycle, where the wake phase is a regular learning phase, meant to acquire new knowledge fast, and the sleep phase is a consolidation phase, where the model reinforces its own knowledge. Although the results presented in \cite{liEnergyBasedModelsContinual2022,singhWakeSleepEnergyBased2024} were promising, their use of backward gradient propagation and synchronous, feed-forward Neural EBM architectures lacked biological plausibility and analog hardware friendliness of Hopfield Networks on which the present work focuses.
Conversely, \cite{kuboLifelongLearningEquilibrium2025} applied the "wake-sleep" cycle to an EqProp-trained Hopfield Network, although their algorithm required the Hopfield Network to transform into a spiking neural network for the sleep phase. 

The closest prior work to ours is \cite{ansSequentialLearningDistributed2004a}, which introduced pseudo-rehearsal using a model and its frozen copy for task solving and replay sample generation, respectively. This paper's approach remains the closest to our own in the literature, although it was tested on small-scale toy problems (with a maximum of 10 examples per task, 2-5 tasks learned sequentially). Their work employed non-EBM recurrent networks trained with Contrastive Hebbian Learning (CHL), an earlier algorithm. In contrast, our EB-Gen method leverages modern Equilibrium Propagation (EqProp), a more powerful and theoretically grounded training paradigm for energy-based models.

\section {Method}\label{sec:method}

We study the behavior of a Hopfield Network trained with Equilibrium
Propagation under non-stationary data. Performance is evaluated on  the incremental versions of the typical image classification benchmarks MNIST and CIFAR10. In
addition to our proposed EB-Gen method, we evaluate two
standard continual learning baselines: Experience Replay (ER) and Generative Replay (Gen Replay). We also use Joint Training (Joint) and FineTune (FT) as upper and lower bounds for the expected average accuracies.

\subsection{Hopfield Network and Equilibrium Propagation}

\paragraph{Hopfield Networks} A Hopfield network is a convergent recurrent neural network that stores information in the
form of attractors of an energy landscape. It consists of $N$ units with
continuous or binary states collected in a state vector $\mathbf{s} =
(s_1,\dots,s_N)^\top$, connected through a symmetric weight matrix $W =
[w_{ij}]_{i,j=1}^N$, with $w_{ij} = w_{ji}$ and $w_{ii} = 0$ (no
self-connections).

The network's dynamics minimize an energy function $E(\mathbf{s},W)$ via a gradient
descent in the state space. This process is called "relaxation" and is formally expressed in equation \ref{eq:hopnet_relax}. In its simplest form, the Hopfield energy is given by equation \ref{eq:hopfield_energy} generally augmented with bias and regularization terms.

\begin{figure}
    \begin{minipage}{0.50\textwidth}
        \begin{equation} \label{eq:hopnet_relax}
            \frac{ds}{dt}
            =
            -\frac{\partial E(\mathbf{s},W)}{\partial s} ,
        \end{equation}
    \end{minipage}
    \hfill
    \begin{minipage}{0.50\textwidth}

        \begin{equation} \label{eq:hopfield_energy}
            E(\mathbf{s},W)
            =
            -\frac{1}{2}
            \sum_{i,j=1}^N
            w_{ij} s_i s_j ,
        \end{equation}
    \end{minipage}
\end{figure}

\noindent Once discretized w.r.t time, the associated relaxation dynamics follows the following equation
\begin{equation}
    s_i[t+1] = \sum_{j=1}^N w_{ij} s_j[t],
\end{equation}

\noindent which closely resembles inference in conventional neural networks.

\paragraph{Equilibrium Propagation}

Equilibrium Propagation is a supervised learning algorithm for Hopfield Networks that
computes weight updates only using local information gathered from two
successive relaxation phases. This eliminates the need for explicit gradient storage or a separate backward pass, making EqProp both biologically and hardware
plausible.

Training proceeds in two phases. In the \emph{free phase}, the input state variables
are \textit{clamped} (fixed) to a data sample $\mathbf{x}$, while the remaining state variables
are left \textit{free} (able to relax). The system relaxes according to the energy dynamics until it
reaches a stable equilibrium state $\mathbf{s}^0$.

In the \emph{nudged phase}, the input remains clamped and a small perturbation
is applied to the output variables by augmenting the energy with a task-related loss.
The network energy becomes

\begin{equation} \label{eq:nudging_energy}
    E^\beta(\mathbf{s},W)
    =
    E(\mathbf{s},W)
    +
    \beta\,\mathcal{L}_{\mathrm{out}}(\mathbf{y},\mathbf{t}),
\end{equation}

\noindent where \(\beta>0\) is a small scalar and $\mathbf{t}$ denotes the output target. The
system then relaxes again to a new equilibrium $\mathbf{s}^\beta$.

The resulting weight update is given by

\begin{equation}
    \Delta W
    \propto
    -\frac{1}{\beta}
    \left(
        \nabla_W E(\mathbf{s}^\beta,W)
        -
        \nabla_W E(\mathbf{s}^0,W)
    \right),
    \label{eq:w_update}
\end{equation}

\noindent which approximates gradient descent on the task-related loss.

\begin{figure}[h!]
    \centering
    \begin{minipage}{0.4\textwidth}
        \centering
        \includegraphics[width=0.5\linewidth]{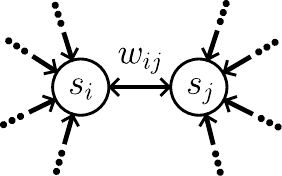}
        \label{fig:bidir_synapse}
    \end{minipage}
    \begin{minipage}{0.30\textwidth}
        \centering
        \begin{equation}
            \Delta w_{ij}
            \propto
            \frac{1}{\beta}
            \big(
                s_i^\beta s_j^\beta
                -
                s_i^0 s_j^0
            \big)
            \label{eq:w_update2}
        \end{equation}
    \end{minipage}
    \hfill

    \caption{The activation values $s_{i\slash j}^{0\slash \beta}$ needed for updating the weight $w_{ij}$ are obtained locally in a topological sense but not in a temporal sense, as they are collected during two different forward phases.}
\end{figure}

For an individual synaptic weight \(w_{ij}\), substituting $E$ in (\ref{eq:w_update}) with the Hopfield energy expression from (\ref{eq:hopfield_energy})
yields equation (\ref{eq:w_update2}) which illustrates the fact that EqProp relies only on local differences between pre- and
post-synaptic activities across the two phases.
These quantities can be accumulated locally, reducing the need for expensive data transfer.

A more complete and faithful description of our Energy-Based Learning setup is provided in annex \ref{sec:ebl_annex}.


\subsection{Problem setting}

\paragraph{Benchmarks}
We choose to study catastrophic forgetting on split-MNIST and Split-CIFAR10. We use the usual partitioning scheme, where classes are grouped following their default order with each group containing the same number of classes.
Split-MNIST and Split-CIFAR10 are composed of 5 task containing 2 class each.
We evaluate those benchmarks in the class-incremental setting, meaning that task identity is never provided at training or test time. Therefore, each output node of our models is permanently associated to a single class and their weights remain modifiable during the training of every task.

\paragraph{Setting limits} Since we focus on class-incremental learning using the Split variations for typical image classification datasets, our analysis is restricted to sequential image classification
problems with clearly defined task boundaries and independently and
identically distributed (i.i.d) samples within each tasks.

\paragraph{Feature extraction}
For experiments on (Split-) CIFAR10, we employ a pre-trained, frozen ResNet-18 feature extractors for the natural image datasets.
Our models will therefore be trained on the latent representations which are generated between the feature extractor and its classification head.
This measure is particularly necessary as hopfield networks, in their basic form, are notoriously difficult to train on datasets which are more challenging than MNIST. 
For the experiments comparing EB-Gen to SRC, we follow the method described by \cite{kuboLifelongLearningEquilibrium2025} and use a truncated VGG13 for Split-CIFAR10, and a ResNet-152 for Split-ImageNet10. 

\paragraph{Incremental learning performance metrics}
We consider a sequence of tasks $T_{1\to T}=(T_1,\dots,T_T)$, where each task
$T_k$ is associated with a dataset $\mathcal{D}^{(k)}$ and learned sequentially.
Let $a_{i,j}$ denote the test accuracy on task $T_j$ after training the model on
tasks $T_1$ through $T_i$. The quantity $a_{i,i}$ measures performance on the
current task immediately after learning $\mathcal{D}^{(i)}$, while $a_{i,j}$
for $j<i$ reflects retention of previously learned tasks.

We summarize performance over the full task sequence using the \textbf{Learning Accuracy (LAcc)}: $\frac{1}{T}\sum_{i=1}^T a_{i,i}$, which reflects plasticity, and the \textbf{Average Accuracy (AAcc)}: $\frac{1}{T}\sum_{i=1}^T a_{T,i}$, which reflects stability.

\subsection{Continual Learning Baselines}
\paragraph{Baselines}
As a frame of reference to evaluate our model's continual performance, we use five baselines: FineTune, ER, Gen Replay, SRC and Joint Training.

FineTune is the most naive strategy where tasks are learned without any forgetting counter-measures, which makes it our lower bound for average accuracy. Conversly, Joint Training refers to the training of a model on the offline task obtained by joining every task together and mixing their samples, making it the upper bound for average accuracy.

We use a simple version of Experience Replay (ER) where samples are selected randomly and the hard predicted labels are used as replay outputs.

For Generative Replay (Gen Replay), we use a separate  MLP-based symmetric VAE generator model to learn input features (for MNIST) or latent features (for CIFAR10) following the method employed by \cite{stoychevLatentGenerativeReplay2023}.

For SRC, we directly use results from \cite{kuboLifelongLearningEquilibrium2025} and compare them to EB-Gen using the same feature extractors for .


\paragraph{General replay principle} The chosen replayed/current sample ratio for a given method is the same for each task as an increasing ratio would imply an explosion of the memory footprint for a higher task count. When using a extracted features, each replay method (ER, Gen Replay and EB-Gen) uses latent features instead of raw input data for buffering or generative training.

\section{Energy-Based Generative Replay (EB-Gen)}
\label{sec:ebgenreplay}

Hopfield Networks trained with EqProp define an explicit energy landscape in which learned information correspond to stable attractors. This allows the same network to be used for both prediction and generation by varying which state variables are constrained during relaxation, enabling the generation of replay samples directly through the network dynamics without auxiliary models or stored training data. 

\subsection{Sample Generation from EqProp-Trained Energy Landscapes}
\label{sec:sample_gen}

After learning a dataset with EqProp, we assume that the energy function
$E(\mathbf{s},W)$ contains multiple stable local minima (i.e. attractors), each capturing some of the data's input-output relationship. While any energy function formally defines a
Boltzmann distribution $p_W(\mathbf{s}) \propto \exp(-E(\mathbf{s},W))$, we do
not directly sample from it. Instead, we generate samples by exploiting the network's relaxation dynamics, which is similar to using a gradient sampling method.

Sample generation in EqProp-trained Hopfield Networks is performed by releasing all state variables and allowing the system to relax to equilibrium. The resulting states $(\mathbf{x}^\ast,\mathbf{y}^\ast)$ correspond to stable attractors of the learned energy landscape and ideally represent previously learned input--output
associations.

However, purely deterministic relaxation yields a limited set of recurrent attractors, reducing sample diversity and coverage of the underlying task distributions. To enhance diversity and improve sample informativeness, we introduce stochasticity during generation—either by injecting noise at each relaxation step (discrete-time Langevin-like dynamics) or by truncating relaxation early, where variability stems from random initialization $s_0$ (typically sampled uniformly over a predefined state range).

The resulting generated configurations provide a diverse set of surrogate
data examples that can be interleaved with new-task data during subsequent training
phases, enabling generative replay without storing past data or introducing
auxiliary generative models.

\subsection{Why Input Reconstruction Must Be Explicitly Trained}

Initial experiments showed that standard EqProp training does not produce models
capable of generating meaningful input samples (see annex \ref{sec:gen_input_ex}). When all state variables are initialized freely
and the system relaxes under the learned energy $E(\mathbf{s},W)$, the resulting
equilibria collapse to simplistic or degenerate input patterns which fail to
capture the structure of the data.

This behavior follows directly from the EqProp training objective. During learning,
input units $\mathbf{x}$ are clamped to observed data, and the task loss
$\mathcal{L}$ depends only on the output variables $\mathbf{y}$. As a result,
the energy landscape is shaped to produce correct outputs conditioned on the
input, but not to assign low energy to input configurations themselves. In
probabilistic terms, EqProp constrains the conditional distribution
$p_W(\mathbf{y},\mathbf{h}\mid\mathbf{x})$ while leaving the marginal
distribution over inputs unconstrained.

When inputs are released during generation, the energy therefore provides no
incentive for $\mathbf{x}$ to converge toward data-like configurations. The
relaxation dynamics can settle in arbitrary input states that remain compatible
with low-energy hidden--output configurations but do not lie on the data
manifold. Explicitly training input reconstruction is therefore necessary to
enable meaningful input generation.

\subsection{Hybrid Training for Input Reconstruction}
\label{sec:input_train}

To improve input generation, we extend the EqProp objective to explicitly train
input reconstruction. During learning, a subset of input units is released and
allowed to evolve under the same relaxation dynamics as the hidden and output
units. Let $\tilde{\mathbf{x}}$ denote the released input variables and
$\mathbf{x}^{\text{data}}$ their corresponding ground-truth values.

We introduce an input reconstruction loss

\begin{equation}
    \mathcal{L}_{\text{in}}(\tilde{\mathbf{x}}, \mathbf{x}^{\text{data}})
    =
    \|\tilde{\mathbf{x}} - \mathbf{x}^{\text{data}}\|^2,
\end{equation}

\noindent which is applied during the nudged phase of EqProp. The perturbed energy
becomes

\begin{equation}
    E^\beta(\mathbf{s},W)
    =
    E(\mathbf{s},W)
    +
    \beta\,
    \Big(
        \mathcal{L}(\mathbf{y}, \mathbf{t})
        +
        \lambda\,\mathcal{L}_{\text{in}}(\tilde{\mathbf{x}}, \mathbf{x}^{\text{data}})
    \Big),
\end{equation}

\noindent where $\lambda$ controls the relative importance of input
reconstruction.

This hybrid objective encourages low-energy equilibria to jointly encode
consistent input--output configurations. As a result, the learned energy
landscape supports meaningful relaxation-based generation of both the inputs (or extracted features for CIFAR10) and outputs of the hopfield network, which is essential for generative replay in continual learning.

\subsection{The Generative Replay Process}
For each new task $T_t$ (with $t>1$), a copy of the network trained on tasks $T_{0\rightarrow t-1}$ is frozen and used to generate new replay samples which are then replayed to the new network alongside new examples. The (replayed/new) batch size ratio is a hyper-parameter of the method, often chosen greater than one. The complete algorithm for EB-Gen is formally described in annex \ref{sec:genreplay_algo}.

\section{Results}
\label{sec:results}

Based on the methodology introduced in Sections~\ref{sec:method} and
\ref{sec:ebgenreplay}, we report results on hybrid EqProp
training, pseudo-example generation, and their use for EB-Gen. The relevant model, training and continual learning hyper parameters for each method are summarized in annex \ref{sec:params}.
\subsection{Offline Hybrid Training}
We first demonstrate the hybrid training of a Hopfield Network with Equilibrium Propagation in the MNIST and CIFAR10 cases.
We evaluate full offline training under three regimes:
\begin{enumerate}
    \item Regular regime: input nodes are fully clamped to input data, no input cost,
    \item Partial regime: Only 90\% of input nodes are clamped to input data, still no input cost,
    \item Hybrid regime: Same as the partial regime, except an input reconstruction cost was added to unclamped input nodes.
\end{enumerate}

\begin{figure}[h]
    \begin{minipage}{0.45\textwidth}
        \centering
        \includegraphics[width=1.0\linewidth]{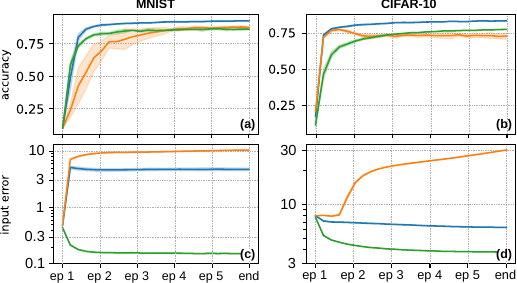}
        \caption{Training performance curves for MNIST and CIFAR10. Averaged over 5 runs.}
        \label{fig:offline_acc}
    \end{minipage}
    \hfill
    \begin{minipage}{0.50\textwidth}

        \centering
        \begin{tabular}{|c|cc|}
            \hline
            Regime  & MNIST & CIFAR10\\
            \hline
            \hline
            Regular & $\textbf{92.5}\pm0.3\%$ & $\textbf{83.5}\pm0.2\%$ \\
            \hline
            Partial & $87.5\pm0.7\%$ & $72.9\pm2.1\%$ \\
            \hline
            Hybrid  & $86.2\pm0.5\%$ & $77.5\pm0.3\%$ \\
            \hline
        \end{tabular}
        \captionof{table}{Final accuracy table}
        \label{tab:offline_acc}
        \vfill
        \begin{tabular}{|c|cc|}
            \hline
            Regime     & MNIST & CIFAR10\\
            \hline
            \hline
            Regular      & $5.5\pm0.2$ & $6.29\pm0.4$ \\
            \hline
            Partial      & $10.5\pm0.0$ & $30.4\pm1.4$\\
            \hline
            Hybrid     &  $\textbf{0.15}\pm0.0$   &  $\textbf{3.80}\pm0.2$ \\
            \hline
        \end{tabular}
        \captionof{table}{Final input reconstruction error table. The training lasted 5 epochs for MNIST and CIFAR10.}
        \label{tab:offline_in_err}
    \end{minipage}
    \hfill

\end{figure}

The evolution of offline accuracy and input error for each regime are shown in figure \ref{fig:offline_acc}, the corresponding final accuracy and input error performances are listed in Table \ref{tab:offline_acc} \& \ref{tab:offline_in_err}. The input errors are computed by freeing 10\% of the model's input nodes and measuring the mean squared error between their equilibrium values and the corresponding input targets. 

While table \ref{tab:offline_acc} shows that using regimes with partial input clamping  substantially degraded accuracy by roughly 5 to 10\%, table \ref{tab:offline_in_err} and figure \ref{fig:offline_acc} show that the hybrid regime successfully improves input reconstruction w.r.t the regular regime. Due to this input/output prediction trade-off, we restrict the hybrid regime to freeing only 10\% of the input nodes for the remainder of the experiments. Notably, the regular regime ended up also reducing reconstruction error for CIFAR-10, while the partial regime saw an explosion of the input error for both datasets, showing that using the input cost is mandatory for enabling input reconstruction.

Our results show that the hybrid training can successfully train our model to improve its partial input reconstruction capability. The next step is to evaluate whether this ability translates to full input generation. 

Examples of image generated using a hopfield network trained on MNIST using the regular and hybrid regime are shown in annex \ref{sec:gen_input_ex}. We observed that, in the hybrid training setting, the generated inputs rarely resembled MNIST digits and often did not even resemble numbers. Whether or not those samples can still be used for performing replay will be assessed in the following section.

\subsection{Continual Learning with Energy-Based Generative Replay}

After showing that EqProp can also be tuned to perform input reconstruction alongside its primary classification task, we evaluate the full EB-Gen method in the incremental setting previously described in section~\ref{sec:method}.

The obtained results are summarized in Table~\ref{tab:aacc_table}, except SRC which is only compared with EB-Gen in Table~\ref{tab:lit_comparison_table}.

\begin{figure}[h]
    \begin{minipage}{0.45\textwidth}
        \begin{center}
            \begin{tabular}{|c|cc|}
                \hline
                Split Datasets      & MNIST & CIFAR10  \\
                \hline
                ER            & $\textbf{85.6}\pm1.0\%$ & $\textbf{76.6}\pm0.5\%$ \\
                Gen Replay    &  $74.8\pm3.2\%$ & $50.5\pm2.7\%$ \\
                EB-Gen     & $75.3\pm1.0\%$ & $56.1\pm0.9\%$ \\
                \hline
                FT            & $19.4\pm0.0\%$ & $21.9\pm0.5\%$ \\
                Joint & $\textbf{91.7}\pm0.6\%$ & $\textbf{82.7}\pm0.2\%$ \\
                Joint-Hybrid & $85.5\pm0.8\%$ & $75.9\pm0.2\%$ \\
                \hline
            \end{tabular}
        \end{center}
        \captionof{table}{Average Accuracy (AAcc) of EqProp-trained models after learning Split-MNIST - CIFAR10 with FT, ER and EB-Gen.}
        \label{tab:aacc_table}
    \end{minipage}
    \hfill
    \begin{minipage}{0.50\textwidth}

        \centering
        \begin{tabular}{|c|c|c|c|c|}
            \hline
            method  &  split dataset  & average accuracy \\
            \hline
              & MNIST & $\textbf{76.6}\pm0.5\%$ \\
            EB-Gen & CIFAR10 & $\textbf{83.4}\pm0.8\%$ \\
              & ImageNet10 & $77.4\%$ \\
            \hline
            FT (ours) & ImageNet10 & $72.6\%$ \\
            \hline
              & MNIST & $64.7\pm 4.35\%$ \\
            SRC  & CIFAR10 & $81.3\pm 1.98\%$ \\
              & ImageNet10 & $\textbf{77.7}\pm 2.91\%$ \\
            \hline
        \end{tabular}
        \captionof{table}{Average Accuracy (AAcc) of EqProp-trained models after learning Split-MNIST - CIFAR10 - ImageNet10 with EB-Gen or SRC. Results are averaged over 5 runs except for ImageNet10.}
        \label{tab:lit_comparison_table}
    \end{minipage}
    \hfill

\end{figure}
The table shows that ER performed the best in terms of stability for both Split MNIST and Split CIFAR10, only losing 5-6\% average accuracy w.r.t to the upper bound defined by Joint. EB-Gen loses at least 10\% average accuracy w.r.t ER, this result is to be linked with the results of the Joint-Hybrid method, in which the joint task is learned using the hybrid training regime, thus losing 6-7\% average accuracy w.r.t regular Joint training. Joint-Hybrid being the actual upper bound for EB-Gen, it partially explains its lower performances. 

Gen Replay performed slightly worst than EB-Gen on both Split-MNIST and Split-CIFAR10.

FineTune performed the worst with 19\%/21\% average accuracy on Split MNIST/CIFAR10, which  corresponds to complete forgetting of four tasks out of five.

\begin{figure}[!h]
    \centering
    \includegraphics[width=1.0\linewidth]{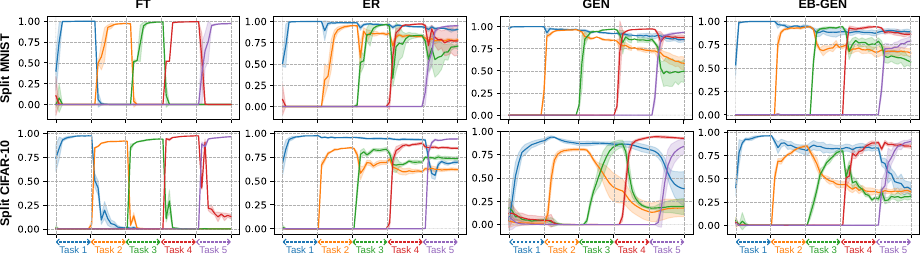}
    \caption{Mean task-wise accuracy evolution during training across four continual learning methods (FT, ER, GEN and EB-Gen) on Split-MNIST.
    Shaded regions indicate one standard deviation across 5 runs.}
    \label{fig:batch_perf_split}
\end{figure}

Figure \ref{fig:batch_perf_split} shows a complete history of the task-specific validation accuracies during training for each continual method on both Split MNIST and Split CIFAR10. 

The FineTune figure shows that each task-specific accuracy instantly drops to zero as soon as the model switches task, indicating absolute and brutal forgetting.
All replay methods show a marked plasticity drop after the first task (especially on Split CIFAR-10), while first-task accuracy remains high throughout. This suggests an over-prioritization of the initial task, impairing performance on later ones—a trend most pronounced in EB-Gen, likely due to its shared parameter space for input and output representations.
We also observe a significant drop of Task 1 accuracy when learning Task 5 with Split CIFAR10, also observable for other task transitions. 


Table \ref{tab:lit_comparison_table} compares the results obtained by ~\cite{kuboLifelongLearningEquilibrium2025} on Split-MNIST - CIFAR10 - ImageNet10) to ours. ImageNet10 is a 10 class subset of Imagenet-1k \cite{russakovsky2015imagenet}. Our method allows a 10\% and a 3\% improvement over SRC on Split-MNIST and Split-CIFAR10.  Results on Split-CIFAR10/ImageNet10 where obtained with a VGG13/ResNet152 feature extractor pre-trained on CIFAR10/ImageNet10 following ~\cite{kuboLifelongLearningEquilibrium2025}'s procedure (whereas our previous CIFAR10 results were obtained with a resNet18 pre-trained on ImageNet-1k). Our updated FT performance shows that pre-training and testing on the same task greatly simplifies the continual learning task, almost making EB-Gen and SRC redundant on Split-ImageNet10 (only gain 5\% average accuracy w.r.t the lower bound).

\section{Analysis}\label{sec:analysis}

\subsection{Forgetting in EBMs} \label{sec:ebm_forgetting}
The forgetting sensitivity of EqProp-trained Hopfield networks (Section \ref{sec:results}) mirrors observations in other EBMs (\cite{liEnergyBasedModelsContinual2022,singhWakeSleepEnergyBased2024}). Below, we formalize a potential explanation for this phenomenon.

A Hopfield Network defines an EBM over state $\mathbf{s}$ with weight matrix $W$, evolving via energy minimization. The energy induces a Boltzmann distribution $p_W(\mathbf{s}) = \frac{1}{Z(W)} \exp\!\left(-E(\mathbf{s}, W)\right)$, where $Z(W)$ is the partition function.

Classical EBM training (e.g., contrastive divergence) approximates maximum likelihood by globally shaping $p_W(\mathbf{s})$ via:

\begin{equation}
    \min_W \;
    \underbrace{
        \mathbb{E}_{\mathbf{s}^+ \sim p_{\text{data}}}
        \big[E(\mathbf{s}^+, W)\big]
    }_{\text{positive phase}}
    \;-\;
    \underbrace{
        \mathbb{E}_{\mathbf{s}^- \sim p_W}
        \big[E(\mathbf{s}^-, W)\big]
    }_\text{negative phase}
\end{equation}

This objective lowers energy for data-supported states while raising it for model-sampled states.

In continual learning with task distributions $\{p_{\text{data}}^{(k)}\}_{k=1}^T$, optimizing for task $k$ shifts probability mass toward $p_{\text{data}}^{(k)}$. Since $p_W$ includes low-energy configurations from prior tasks $j<k$, the negative phase inadvertently raises their energy, eroding earlier minima and causing forgetting.

EqProp resembles contrastive divergence: the free phase captures the model’s prior via relaxation, while the nudged phase biases equilibrium toward data. Thus, the above analysis applies.

\subsection{Replay-based Continual Learning, Memory Footprint and Analog Hardware Friendliness}
Despite underperforming ER on Split-MNIST-CIFAR10 and conventional Gen Replay on Split-MNIST (Section \ref{sec:results}), EB-Gen retains key advantages: its Hybrid regime uses standard EqProp for input generation, and replay data emerges solely from the Hopfield network’s natural dynamics. This eliminates auxiliary modules, making EB-Gen far more hardware-efficient—a core benefit of EqProp.

In contrast, ER would require either external digital memory or on-chip capacitive arrays (likely consuming significant chip area), while Gen Replay necessitates a separate generative model, incurring costs for training, inference, and weight storage.

\section{Conclusion}

 We propose EB-Gen, a continual learning framework that enables EqProp-trained Hopfield Networks to  learn in a class-incremental setting without sacrificing their appeal to analog hardware deep learning. Our approach employs the underlying Hopfield Network both as a classifier for learning new tasks and a generator for remembering old ones, thus alleviating the need for additional memory or generator modules. We demonstrated that our method helps significantly reduce catastrophic forgetting, confirming that EqProp-trained Hopfield Networks can effectively retain meaningful information across both input and output data spaces.confirming that EqProp-trained Hopfield Networks can effectively retain meaningful information across both input and output data spaces. Moreover, since EB-Gen encodes past data within the network’s weights and produces non–human-readable samples (see annex \ref{sec:gen_input_ex}), it may offer an opportunity for enhanced data privacy.

\section{Future work} In the future, we would like to replicate the presented results on a greater variety of continual learning scenarios, such as Domain-Incremental scenarios or task-boundary-less scenarios.  
Additionally, we aim to broaden the comparison with the continual learning state of the art, including parameter regularisation or isolation-based methods.
Additionally, we believe that there is room for improving the current version of Energy-Based GenReplay, by testing other weight topologies or trying out typical generative replay tricks, such non-trivial sample selection strategies.

\bibliography{collas2026_conference}
\bibliographystyle{collas2026_conference}

\newpage
\appendix
\section{Exact Energy Based Learning Setup} \label{sec:ebl_annex}
\subsection{Hopfield Network}
For our experiments, we used the same Hopfield energy formulation as the origninal EqProp paper by \cite{scellierEquilibriumPropagationBridging2017a}. 
Given a state $S=(s_1,\dots,s_N)$ and a weight matrix $W=(w_{ij})_{i,j=1}^{N}$, the Hopfield energy takes the following form:
\begin{equation} \label{eq:true_hop_energy}
    E(S,W)=-\frac 1 2\sum_i^N\sum_{j\neq i}^N \rho(s_i) w_{ij} \rho(s_j) -\sum_i b_i \rho(s_i) +\frac 1 2 ||S||^2
\end{equation}
Where $-\frac 1 2\sum_i^N\sum_{j\neq i}^N \rho(s_i) w_{ij} \rho(s_j)$ represents the energy of the interaction between units (with $\rho$ an activation function) and $-\sum_i b_i s_i$ is the energy associated to the individual biases. 
The general relaxation equation stays the same, with the optional noise for generation:
\begin{equation}
    \frac{dS}{dt} = -\nabla_S E +\sqrt{2}\frac{dW}{dt}
\end{equation}
With $\nabla_S E$ the energy state gradient and $W$ a wiener process (a Gaussian noise with variance $dt$).
The time-discretized version of this equation for a singular state unit is $s_i$ is $$s_i^{t+1}=s_i^t-\Delta t \frac{\partial E}{\partial s_i} + \sqrt{2}\Delta t z_i^t$$ with $z_i^t \sim \mathcal{N}(0,1)$.
Plugging the Hopfield Energy in the previous formula yields 
$$s_i^{t+1}=s_i^t(1-\Delta t)+\Delta t\: \rho'(s_i) \big(\sum_{j\neq i}^N  w_{ij} \rho(s_j)+b_i\big) + \sqrt{2}\Delta t z_i^t$$
where $\Delta t\: \rho'(s_i) \big(\sum_{j\neq i}^N  w_{ij} \rho(s_j)+b_i\big)$ is the interaction term, $\sqrt{2}\Delta t z_i^t$ is the noise term, and $s_i^t(1-\Delta t)$ is a state leakage term.

\subsection{Equilibrium Propagation}
We use a variant of the original EqProp formulation introduced by ~\cite{laborieuxScalingEquilibriumPropagation2021} where a third 'negative' nudging phase is computed after the usual 'positive' nudging phase in order to better approximate the true loss gradient. 
During the negative nudging phase, equation \ref{eq:nudging_energy} becomes
\begin{equation}
    E^\beta(\mathbf{s},W)
    =
    E(\mathbf{s},W)
    -
    \beta\,\mathcal{L}_{\mathrm{out}}(\mathbf{y},\mathbf{t}),
\end{equation}
with the only difference being the sign of $\beta$. The equilibrium state retrieved from this phase is $s^{-\beta}$, it corresponds to a state which is particularly bad candidate for correctly associating the input image and the corresponding true target. 

The resulting weight update becomes

\begin{equation}
    \Delta W
    \propto
    -\frac{1}{2\beta}
    \left(
        \nabla_W E(\mathbf{s}^\beta,W)
        -
        \nabla_W E(\mathbf{s}^{-\beta},W)
    \right),
    \label{eq:centered_w_update}
\end{equation}

Since, in the case of EB-Gen, we also partially learn the input images, we had to choose whether the input cost beta factor should also become negative during the negative nudging phase. Since the goal of the negative phase is to ensure that the model doesn't associate the write inputs with the wrong outputs, we kept a positive beta value for the inputs. Trials of both negative and positive beta values proved that it was the most efficient choice. 

Plugging equation \ref{eq:true_hop_energy} into equation \ref{eq:centered_w_update} for a specific weight $w_{ij}$ yields
\begin{equation}
    \Delta w_{ij}
    \propto
    -\frac{1}{2\beta}
    \left(
        \rho(\mathbf{s_i}^\beta) \rho(\mathbf{s_j}^\beta)
        -
        \rho(\mathbf{s_i}^{-\beta}) \rho(\mathbf{s_j}^{-\beta})
    \right),
    \label{eq:centered_w_update_local}
\end{equation}
which shows that the centered version of EqProp doesn't change the locality of the weight update.

\newpage
\section{Pseudo-Example Generation with EqProp-trained Hopfield Networks}

\subsection{Generated Input Examples} \label{sec:gen_input_ex}
We exclusively conduct the evaluation on MNIST as the models trained on CIFAR10 use feature extractors, they can therefore only regenerate latent features which are likely to not be human-readable by default. Examples are generated with networks trained  under four different regimes:

\begin{itemize}
    \item an untrained network,
    \item training with full data input,
    \item training with $1\%$ input damage and an input reconstruction cost,
    \item training with $10\%$ input damage and an input reconstruction cost.
\end{itemize}

\begin{figure}[!h]
    \centering
    \includegraphics[width=0.8\linewidth]{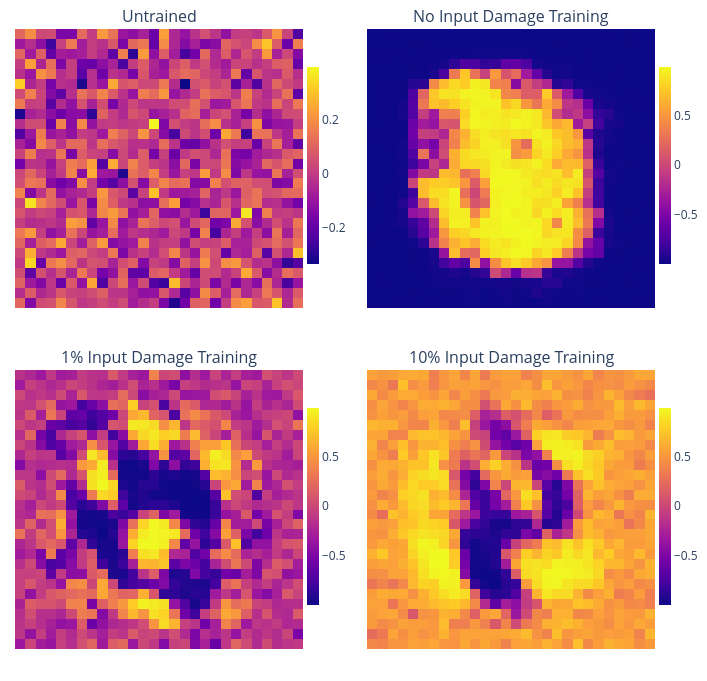}
    \caption{Input samples generated after full input release for different
    training regimes.}
    \label{fig:100p_in_reco}
\end{figure}
When trained without input reconstruction, the network relaxes toward a
diffuse, centered activation pattern that only captures coarse properties of
MNIST, such as spatial extent and centering. While this reflects some dataset
statistics, it provides no meaningful digit structure.

In contrast, models trained with partial input release and an explicit input
reconstruction cost produce significantly more structured patterns. Although
these patterns do not resemble human-interpretable images, this is not
necessarily problematic: deep models often encode useful information in
non-human-readable representations (such as filters used in CNNs). To assess
whether these generated patterns carry task-relevant information, we need to evaluate their effectiveness for knowledge transfer, which we do in an online setting when testing EB-Gen.
\newpage

\subsection{Generative Replay Algorithm} \label{sec:genreplay_algo}

Algorithm \ref{alg:ep_ebgr} details the full functioning of EB-Gen, including the generation and training part.
Tasks are learned sequentially using EqProp on batches that
combine current-task data with pseudo-samples generated by a frozen copy of the model. Replay is implemented directly through the relaxation dynamics of the learned energy function, without storing past data or introducing an additional independent generative models.

\begin{algorithm}[!t]
\caption{Continual Learning with Equilibrium Propagation and EB-Gen}
\label{alg:ep_ebgr}
\begin{algorithmic}[1]

\Require Sequence of tasks $\{\mathcal{D}^{(k)}\}_{k=1}^T$, where
$\mathcal{D}^{(k)} = \{(\mathbf{x}, \mathbf{t})\}$ and
$\mathbf{t} \in \{0,1\}^C$.

\Require EqProp hyperparameters:
$\beta$, learning rate $\eta$, step size $\Delta t$, time constant $\tau$,
free/nudged steps $K_{\mathrm{free}},K_{\mathrm{nudged}}$, tolerance $\epsilon$.

\Require Replay hyperparameters:
pseudo ratio $r$, mask fraction $\rho$, noise scale $\sigma$,
sampling steps $K_{\mathrm{gen}}$.

\Require Hopfield energy: $E(\mathbf{s},W)$ over $\mathbf{s}=(\mathbf{x},\mathbf{h},\mathbf{y})$. Losses:
$\mathcal{L}_{\mathrm{out}}(\mathbf{y},\mathbf{t})$,
$\mathcal{L}_{\mathrm{in}}(\tilde{\mathbf{x}},\mathbf{x})$, weight $\lambda$.

\vspace{0.5em}
\State \textbf{Initialize parameters} $W \sim \mathrm{Init}(\cdot)$
\vspace{0.5em}

\For{$k = 1,\dots,T$} \Comment{Iterate over tasks}

    \State Set teacher $W_{\mathrm{teach}} \gets W$ and freeze it

    \For{each minibatch $\mathcal{B}=\{(\mathbf{x}_b,\mathbf{t}_b)\}_{b=1}^B$ from $\mathcal{D}^{(k)}$}

        \Statex
        \Statex \textbf{(A) Pseudo-example generation (teacher model)}
        \State Initialize $\mathcal{B}_{\mathrm{ps}} \gets \emptyset$

        \For{$m = 1,\dots,rB$}
            \State Initialize state $\mathbf{s}_0 \sim \mathrm{Unif}(\text{state range})$
            \State Release all state variables $(\mathbf{x},\mathbf{h},\mathbf{y})$

            \For{$t=0,\dots,K_{\mathrm{gen}}-1$}
                \State $\mathbf{s}_{t+1} \gets \mathbf{s}_t
                - \frac{\Delta t}{\tau}\nabla_{\mathbf{s}}E(\mathbf{s}_t,W_{\mathrm{teach}})
                + \sqrt{\frac{2\Delta t}{\tau}}\sigma\,\boldsymbol{\xi}_t$
                \Comment{$\boldsymbol{\xi}_t \sim \mathcal{N}(0,I)$}
            \EndFor

            \State Extract $(\mathbf{x}^{\mathrm{ps}},\mathbf{y}^{\mathrm{ps}})$ from $\mathbf{s}_{K_{\mathrm{gen}}}$
            \State Add $(\mathbf{x}^{\mathrm{ps}},\mathbf{y}^{\mathrm{ps}})$ to $\mathcal{B}_{\mathrm{ps}}$
        \EndFor

        \State Form mixed batch $\mathcal{B}_{\mathrm{mix}} \gets \mathcal{B} \cup \mathcal{B}_{\mathrm{ps}}$
        \Statex
        \Statex \textbf{(B) EqProp training on mixed batch}
        \State Initialize accumulated update $\Delta W \gets 0$

        \For{each $(\mathbf{x},\mathbf{t}) \in \mathcal{B}_{\mathrm{mix}}$}

            \State Sample mask $M$ and define $\mathbf{x}^{\mathrm{cl}}$
            \State Initialize state $\mathbf{s}$ and clamp $\mathbf{x}^{\mathrm{cl}}$

            \Statex \emph{Free phase}
            \For{$i=1,\dots,K_{\mathrm{free}}$}
                \State $\mathbf{s} \gets \mathbf{s}
                - \frac{\Delta t}{\tau}\nabla_{\mathbf{s}}E(\mathbf{s},W)$
            \EndFor
            \State Store $\mathbf{s}^0 \gets \mathbf{s}$

            \Statex \emph{Nudged phase}
            \State Define
            $E^\beta = E + \beta\big(\mathcal{L} + \lambda\mathcal{L}_{\mathrm{in}}\big)$
            \For{$i=1,\dots,K_{\mathrm{nudged}}$}
                \State $\mathbf{s} \gets \mathbf{s}
                - \frac{\Delta t}{\tau}\nabla_{\mathbf{s}}E^\beta(\mathbf{s},W)$
            \EndFor
            \State Store $\mathbf{s}^\beta \gets \mathbf{s}$

            \State Accumulate
            $\Delta W \gets \Delta W
            - \frac{1}{\beta}\big(\nabla_W E(\mathbf{s}^\beta,W)
            - \nabla_W E(\mathbf{s}^0,W)\big)$
        \EndFor

        \State Update parameters
        $W \gets W + \eta\,\Delta W / |\mathcal{B}_{\mathrm{mix}}|$

    \EndFor
\EndFor
\end{algorithmic}
\end{algorithm}

\newpage
\section{Experiment Parameters}\label{sec:params}
\begin{table}[htbp]
    \begin{center}
        \begin{tabular}{|c|ccc|}
            \hline
             Parameters    & MNIST & CIFAR10 & ImageNet10\\
            \hline
            \hline
            $lr1$     & 0.01 & 0.001 & 0.08 \\
            $lr2$     & 0.01 & 0.001 & 0.001\\
            $\beta$    & 1.0 & 1.0 & 1.0  \\
            \hline
            $dt$    & 0.1 & 0.1 & 0.1 \\
            relax tol & 1e-3 & 1e-3  & 1e-3  \\
            free phase & 250 & 500   & 500  \\
            nudge phase & 50 & 50 & 50 \\
            \hline
            Hidden size & 125 & 512 & 1024 \\
            extractor & None & ResNet18 & ResNet152 \\
            \hline
            batch size & 256 & 256 & 256 \\
            epoch per task & 2 & 2  & 5 \\
            optim & Adam & Adam & Adam \\
            loss & MSE & MSE & MSE \\
            \hline
        \end{tabular}
    \end{center}
    \caption{\textbf{General Hyper-Parameters} for each tested dataset. \textbf{Relaxation} $dt$ is the discrete time step size for the energy relaxation; 'relax tol' is the relaxation tolerance, where relaxation stops if the state update amplitude goes below it; 'free phase' and 'nudge phase' are the maximum number of steps for the free phase and nudged phase relaxations respectively. \textbf{Model} 'Hidden size' is the size of the network's hidden node; $lr1$ and $lr2$ are the learning rates for the first and second weight layers; "extractor" is the name of the used frozen feature extractor. \textbf{Training} 'optim' is the optimizer used to train the main model, the same 'loss' is used for output and input nodes. For the comparison with SRC, the CIFAR10 parameters where the same except the efature extractor, which was a truncated VGG13 (with batch norm)}
\end{table}

\begin{table}[h!]
    \begin{center}
        \begin{tabular}{|c|cc|}
            \hline
             Parameters    & MNIST & CIFAR10  \\
            \hline
            Buffer size & 128 & 512 \\
            Replay ratio & 1 & 2\\ 
            \hline
        \end{tabular}
    \end{center}
    \caption{\textbf{Experience Replay}-specific hyper parameters. 'Buffer size' is the maximum number of data samples ER can keep at a time. 'Replay ratio' is the replay/current sample size ratio used for a single training step.}
\end{table}

\begin{table}[h!]
    \begin{center}
        \begin{tabular}{|c|cc|}
            \hline
             Parameters    & MNIST & CIFAR10  \\
            \hline
            $lr1$     & 0.01 & 0.005 \\
            $lr2$     & 0.01 & 0.005 \\
            latent size & 128 & 128 \\
            Replay ratio list & 1,2,2,2 & 1,2,2,2 \\
            gen lr & 0.01 & 0.005 \\
            \hline
        \end{tabular}
    \end{center}
    \caption{\textbf{Generative Replay}-specific hyper parameters. WWe chose special $lr1$ and $lr2$ values for training Gen Replay on CIFAR10. 'Replay ratio list' specifies the replay/current sample size ratio associated to each task of the task sequence, starting from the second task (first one doesn't need replay). 'latent size' is the size of the latent space of the VAE. 'gen lr' is the learning rate used for training the generation model. The rest of the generator architecture and training parameters follows the one described by \cite{stoychevLatentGenerativeReplay2023}. }
    \label{tab:gen_hp}
\end{table}

\begin{table}[h!]
    \begin{center}
        \begin{tabular}{|c|ccc|}
            \hline
             Parameters & MNIST & CIFAR10 & ImageNet10 \\
            \hline
            Replay ratio list & 0.5,1,2,4 & 1,4,4,4 & 0.5,0.5,0.5,0.5\\
            input loss & MSE & MSE & MSE\\
            gen phase & 100 & 200 & 200 \\
            chain len & 10 & 20 & 100\\
            \hline
        \end{tabular}
    \end{center}
    \caption{\textbf{Energy Based Generative Replay}-specific hyper parameters. 'Replay ratio list' has the same meaning here as in Table \ref{tab:gen_hp}. 'gen phase' is the maximum number of steps the generation process can take; 'chain len' is the amount of generated samples which are retrieved for a single relaxation.}
\end{table}

\end{document}